\documentclass{article}

\usepackage[nonatbib, preprint]{neurips_2020}

\usepackage[utf8]{inputenc} 
\usepackage[T1]{fontenc}    
\usepackage{hyperref}       
\usepackage{url}            
\usepackage{booktabs}       
\usepackage{amsfonts}       
\usepackage{nicefrac}       
\usepackage{microtype}      
\usepackage{graphicx}       
\usepackage{color, colortbl} 
\usepackage{multirow} 
\usepackage{longtable}
\usepackage[most]{tcolorbox} 
\usepackage{lipsum} 
\usepackage[utf8]{inputenc}
\usepackage{tcolorbox}
\usepackage{amsmath}
\usepackage{tcolorbox}

\usepackage{booktabs} 
\usepackage{tabularx} 
\usepackage{geometry} 

\usepackage[utf8]{inputenc}
\usepackage{enumitem}

\definecolor{headercolor}{gray}{0.85}
\definecolor{maxvalue}{rgb}{0.92,1,0.92} 

\tcbuselibrary{skins}

\title{IMAS: A Comprehensive Agentic Approach to Rural Healthcare Delivery}

\author{%
  Agasthya Gangavarapu\\
Founder, uheal.ai\\
  \texttt{august@uheal.ai} \\
   \And
  Ananya Gangavarapu \\
  AI Engineer, Ethriva \\
  \texttt{ananya@ethriva.ai} \\
}

\begin{document}

\maketitle

\begin{abstract}
Since the onset of COVID-19, rural communities worldwide have faced significant challenges in accessing healthcare due to the migration of experienced medical professionals to urban centers. Semi-trained caregivers, such as Community Health Workers (CHWs) and Registered Medical Practitioners (RMPs), have stepped in to fill this gap, but often lack formal training. This paper proposes an advanced agentic medical assistant system designed to improve healthcare delivery in rural areas by utilizing Large Language Models (LLMs) and agentic approaches. The system is composed of five crucial components: translation, medical complexity assessment, expert network integration, final medical advice generation, and response simplification. Our innovative framework ensures context-sensitive, adaptive, and reliable medical assistance, capable of clinical triaging, diagnostics, and identifying cases requiring specialist intervention. The system is designed to handle cultural nuances and varying literacy levels, providing clear and actionable medical advice in local languages. Evaluation results using the MedQA, PubMedQA, and JAMA datasets demonstrate that this integrated approach significantly enhances the effectiveness of rural healthcare workers, making healthcare more accessible and understandable for underserved populations. All code and supplemental materials associated with the paper and IMAS are available at  \textit{https://github.com/uheal/imas}.
\end{abstract}

\section{Introduction}

Since the onset of COVID-19, rural communities around the world have encountered significant challenges in accessing healthcare. Financial incentives have increasingly drawn qualified and experienced medical professionals to urban centers, leaving rural areas underserved. To address this gap, semi-trained caregivers, such as Registered Medical Practitioners (RMP) in India \cite{ncta_course_215}, and Community Health Workers (CHWs) are stepping in to provide essential healthcare services. However, these caregivers often operate with limited or no formal training, which poses further challenges to the quality and effectiveness of healthcare in rural settings. Large Language Models (LLMs) and agentic approaches, when applied to the healthcare domain, have the potential to be valuable tools in various healthcare areas. These technologies can provide contextual medical training, assist in diagnostics, and support the treatment of various simple to moderately complex clinical cases, particularly for CHWs and other healthcare workers in rural communities.

To make agentic medical assistants effective in rural parts of the world, these assistants should be context-sensitive, adaptive, and reliable. While foundational models such as GPT-4 and Llama 3 have been proven effective when evaluated using medical benchmarks like PubMedQA \cite{jin2019pubmedqa}, their performance in multi-turn medical dialogs and medical decision-making is limited and requires significant engineering. Additionally, these assisting systems must be capable of clinical triaging and identifying complex cases that require specialist intervention and referral to a regional health center. Furthermore, the cultural nuances and limited literacy prevalent among rural health workers and patients make standard medical responses difficult to understand and act upon. For example, the Telugu term \textit{Vedi cheyatam}, a common health condition term used by more than 100 million people \cite{wikipedia_telugu_people}, has no direct English equivalent. Most models return out-of-context meanings and unrelated diagnoses when used in the list of symptoms.

This paper proposes an agentic medical assistant system that leverages domain-adopted Large Language Models (LLMs) to provide context-sensitive medical triaging and diagnostics. The system contains five steps: 1) Translation; 2) Medical Complexity Check; 3) Leverage Expert Network; 4) Final medical advice and 5) Simplify and Filter.

\section{Related Works}

LLMs are being increasingly customized for a range of applications within the healthcare and medical fields. These domain-adapted models are performing at human expert levels in tasks such as question answering, reading and generating medical reports, and clinical diagnosis. Most of these models are built predominantly using two approaches: 1) fine-tuning the foundational model with domain-specific datasets, and 2) using prompting and retriever systems. While these techniques have shown remarkable improvements in pretrained LLMs, they still perform unreliably in healthcare settings in rural areas.

Our approach leverages fine-tuned LLMs in combination with multidisciplinary, contextual, and collaborative agents to enhance the performance and reliability of the healthcare system. While existing agent collaboration models incorporate techniques like role-playing, group discussions, and negotiation, I utilize the MDAgent framework\cite{MDAgents2024}, which dynamically selects the optimal collaboration strategy for execution.

\section{IMAS: Integrated Medical Agent System}

The Medical Assistant System (MAS) is designed to leverage best-of-breed architectural approaches to deliver context-sensitive medical advice in local languages. The key design principles are as follows:
\begin{itemize}
    \item \textbf{Local Language Interaction}: Rural healthcare workers, such as RMPs, can interact with the system in their local language and receive assistance in the same language.
    \item \textbf{Sensitivity to Local Context}: The system should be sensitive to the local population's literacy levels and understanding to ensure they can act on the medical advice provided.
    \item \textbf{Guardrails for Misinformation}: The system should include guardrails to prevent misinformation, disinformation, and toxic prompting.
    \item \textbf{Privacy and Security}: The privacy and security of all participants, especially patients and health workers, should be protected while interacting with the system.
     \item \textbf{Accessibility}: The system should be accessible on commonly used smart mobile devices.
\end{itemize}

The system consists of the following key components:

\begin{figure}[ht]
    \centering
    \includegraphics[width=\linewidth,height=3in]{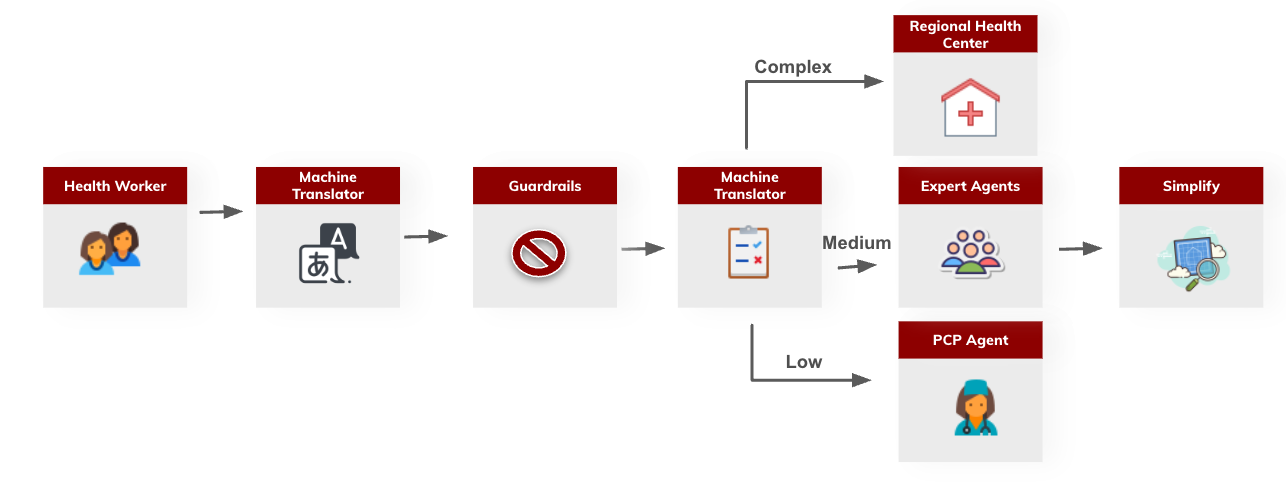} 
    \caption{High-level usage flow of IMAS}
    \label{fig:example}
\end{figure}

\begin{itemize}
    \item \textbf{Medical Large Model}: Llama 3 70B model fine-tuned with medical dialog datasets to enhance its performance in healthcare applications.
    \item \textbf{Translation Model}: Meta's open-sourced \textit{Seamless-M4t-v2-large} \cite{seamless_communication} is fine-tuned with local language and vernacular terminology.
    \item \textbf{Collection of Agents}: These agents collectively evaluate the complexity of medical cases, perform diagnostic procedures adaptively, simplify medical responses for better understanding, orchestrate system interactions, and ensure responses are empathetic and patient-centric.
    \item \textbf{Integrated Guardrail System}: To mitigate hallucinations and misinformation, the system incorporates an integrated guardrail framework using \textit{NeMo Guardrails} \cite{nemo_guardrails}and \textit{Llama Guard}\cite{dubey2024llama3}, ensuring safer, more reliable interactions across medical applications.
\end{itemize}

\subsection{Agents and Their Roles}

\textbf{Complexity Assessment Agent}: This agent functions as a general practitioner (GP) who evaluates the complexity of each medical case presented. If the case of high complexity, then the case is referred to formally trained doctor probably in Regional Healthcare Center (RHC). Medical complex assessment criteria is at high-level:
\begin{itemize}
    \item \textbf{Low Complexity Cases}: These involve simple, well-defined medical issues that can be resolved typically by a single Primary Care Provider (PCP).  Examples include minor infections, routine check-ups, and ongoing management of well-controlled diabetes or hypertension.
    \item \textbf{Medium Complexity Cases}: These involve medical issues with multiple interacting factors, necessitating a collaborative approach among a multidisciplinary team (MDT). Examples include managing patients with coexisting chronic conditions, complex diagnostic cases, and treatments requiring the expertise of various medical specialists.
    \item \textbf{High Complexity Cases}: These involve highly complex medical scenarios that demand extensive coordination and combined expertise from an Integrated Care Team (ICT). Examples include managing patients with severe multi-organ diseases, intricate post-surgical care, and comprehensive treatment plans for complex traumas.
\end{itemize}

\textbf{Collaborative Diagnostic Agents}: A set of expert agents that work together adaptively to perform diagnostic procedures. Based on the Complexity Assessment Agent's evaluation, these agents handle low and medium-complexity cases only. They leverage Electronic Medical Records (EMR) and expert knowledge to provide accurate diagnoses and recommend appropriate treatment plans. The collaborative nature of these agents ensures that multiple perspectives and expertise are utilized, improving the accuracy and reliability of the diagnostic process.\\

\textbf{Response Simplification Agents}: These agents simplify the medical responses to make them understandable for patients and healthcare workers. They take the complex medical jargon and technical details provided by the diagnostic agents and translate them into clear, concise, and actionable information. Additionally, they incorporate important guardrails to ensure the accuracy and safety of the information provided. Their functions include:
\begin{itemize}
    \item Language Simplification: Breaking down complex medical terminology into simple language that can be easily understood by individuals with varying levels of literacy.
    \item Cultural Adaptation: Tailoring the responses to be culturally relevant and sensitive to the local context, ensuring that the advice is not only understandable but also acceptable and actionable within the patient's cultural framework.
    \item Step-by-Step Instructions: Providing detailed, step-by-step instructions for any recommended actions or treatments, making it easier for patients and healthcare workers to follow the advice correctly.
    \item Guardrails for Misinformation: Implementing safeguards to detect and prevent misinformation, disinformation, and toxic prompting, ensuring that the information provided is accurate, reliable, and safe for patients and healthcare workers.
\end{itemize}

\section{Results}
To evaluate the IMAS, multiple benchmarks are being used. Traditional benchmarks such as MedQA and PubMedQA are employed alongside more specific datasets like guidelines. These datasets have been translated from various languages, primarily Chinese and English, into Telugu, Hindi, Swahili, and Arabic using Azure Cognitive Services.

\begin{table}[htbp]
\centering

\begin{tabularx}{\textwidth}{@{}Xccccc@{}}
\toprule
\textbf{Model} & \multicolumn{3}{c}{\textbf{Knowledge Retrieval}} & \multicolumn{2}{c}{\textbf{Diagnostic Tasks}} \\
\cmidrule(lr){2-4} \cmidrule(lr){5-6}
& MedQA & PubMedQA & Guidelines & JAMA & DDXPlus \\
\midrule
GPT-4* & 79.7 & 67.2 & 64.3 & - & - \\
Llama-3 70B* & 78.2 & 67.5 & 61.8 & & \\
\midrule
MedAgents* & 79.1 & 69.7 & 54.6 & 66.0 & 62.8 \\
MdAgents* & \cellcolor{maxvalue}\textbf{88.7} & \cellcolor{maxvalue}\textbf{75.0} & 65.3 & \cellcolor{maxvalue}\textbf{70.9}& \cellcolor{maxvalue}\textbf{77.9}\\
\textbf{IMAS} & 78.9 &74.1 & \cellcolor{maxvalue}\textbf{66.8}&68.9 & 76.8\\
\bottomrule
\end{tabularx}
\vspace{6pt}
\caption{Evaluation results of high-performing foundational models, agentic systems, and IMAS. \\ *Extracted from system cards of the respective models \cite{OpenAI2023GPT4, Meta2024Llama3, MDAgents2024, tang2023MedAgents} and may not be a true representation of the performance with translated benchmark data.}
\label{tab:performance}
\end{table}
With benchmarking datasets such as MedQA, PubMedQA \cite{jin2019pubmedqa}, and JAMA \cite{chen2024benchmarking}, the emphasis is on evaluating models and agents in synthesizing various aspects of medical knowledge. Similarly, diagnostic tasks like DDXPlus \cite{tchango2022ddxplus} assess the ability of models to systematically analyze inputs and generate accurate diagnoses. The proposed framework demonstrated competitive performance, comparable to state-of-the-art models, when employing six specialized agents. However, reducing the number of agents to four, each focusing on different domains, resulted in a performance drop of over 10\%. While replacing the fine-tuned custom Llama 3 70B model with GPT-4 led to an overall performance boost, results varied significantly across languages. Notably, Telugu and Hindi showed stronger performance compared to Arabic and Swahili. The overall system performance improved considerably when different models were assigned to different agents, particularly benefiting the Response Simplification Agent. Specific details regarding GPT-4's performance and the optimal number of agents in the framework still require further clarification.

\section{Conclusion}
This project introduces an agent framework designed to enhance the application of LLMs as medical assistants for rural healthcare workers. The Integrated Medical Assistant System (IMAS) systematically evaluates cases, adaptively makes diagnostic decisions, and delivers these decisions in a culturally sensitive manner. The system demonstrates performance that ranks near the top on all relevant medical benchmarks.

\clearpage
\medskip

\small


\begin{thebibliography}{10}

\bibitem{Meta2024Llama3}
Meta AI.
\newblock Llama 3 system card.
\newblock \url{https://ai.facebook.com/research/llama-3}, 2024.
\newblock Accessed: 2024-06-27.

\bibitem{chen2024benchmarking}
Hanjie Chen, Zhouxiang Fang, Yash Singla, and Mark Dredze.
\newblock Benchmarking large language models on answering and explaining challenging medical questions, 2024.
\newblock arXiv.org perpetual non-exclusive license.

\bibitem{wikipedia_telugu_people}
Wikipedia contributors.
\newblock Telugu people — wikipedia, the free encyclopedia.
\newblock \url{https://en.wikipedia.org/wiki/Telugu_people}, 2024.
\newblock Accessed: 2024-10-08.

\bibitem{dubey2024llama3}
Abhimanyu Dubey, Abhinav Jauhri, Abhinav Pandey, Abhishek Kadian, Ahmad Al-Dahle, Aiesha Letman, Akhil Mathur, Alan Schelten, Amy Yang, Angela Fan, Anirudh Goyal, Anthony Hartshorn, Aobo Yang, Archi Mitra, Archie Sravankumar, Artem Korenev, Arthur Hinsvark, Arun Rao, Aston Zhang, Aurelien Rodriguez, Austen Gregerson, Ava Spataru, Baptiste Roziere, Bethany Biron, Binh Tang, Bobbie Chern, Charlotte Caucheteux, Chaya Nayak, Chloe Bi, Chris Marra, Chris McConnell, Christian Keller, Christophe Touret, Chunyang Wu, Corinne Wong, Cristian~Canton Ferrer, Cyrus Nikolaidis, Damien Allonsius, Daniel Song, Danielle Pintz, Danny Livshits, David Esiobu, Dhruv Choudhary, Dhruv Mahajan, Diego Garcia-Olano, Diego Perino, Dieuwke Hupkes, Egor Lakomkin, Ehab AlBadawy, Elina Lobanova, Emily Dinan, Eric~Michael Smith, Filip Radenovic, Frank Zhang, Gabriel Synnaeve, Gabrielle Lee, Georgia~Lewis Anderson, Graeme Nail, Gregoire Mialon, Guan Pang, Guillem Cucurell, Hailey Nguyen, Hannah Korevaar, Hu~Xu, Hugo Touvron, Iliyan Zarov,
  Imanol~Arrieta Ibarra, Isabel Kloumann, Ishan Misra, Ivan Evtimov, Jade Copet, Jaewon Lee, Jan Geffert, Jana Vranes, Jason Park, Jay Mahadeokar, Jeet Shah, Jelmer van~der Linde, Jennifer Billock, Jenny Hong, Jenya Lee, Jeremy Fu, Jianfeng Chi, Jianyu Huang, Jiawen Liu, Jie Wang, Jiecao Yu, Joanna Bitton, Joe Spisak, Jongsoo Park, Joseph Rocca, Joshua Johnstun, Joshua Saxe, Junteng Jia, Kalyan~Vasuden Alwala, Kartikeya Upasani, Kate Plawiak, Ke~Li, Kenneth Heafield, and Kevin~Stone et~al. (434 additional authors~not shown).
\newblock The llama 3 herd of models.
\newblock {\em arXiv preprint arXiv:2407.21783}, 2024.

\bibitem{ncta_course_215}
National~Centre for Technical Assistance~(NCTA).
\newblock Course on {Advanced AI Applications in Healthcare}.
\newblock \url{https://nctaindia.in/courses?course_id=215}, 2024.
\newblock Accessed: 2024-10-08.

\bibitem{jin2019pubmedqa}
Qiao Jin, Bhuwan Dhingra, Zhengping Liu, William Cohen, and Xinghua Lu.
\newblock Pubmedqa: A dataset for biomedical research question answering.
\newblock In {\em Proceedings of the 2019 Conference on Empirical Methods in Natural Language Processing and the 9th International Joint Conference on Natural Language Processing (EMNLP-IJCNLP)}, 2019.

\bibitem{MDAgents2024}
MIT~Media Lab.
\newblock Mdagents: Multidisciplinary agents framework.
\newblock \url{https://github.com/mitmedialab/MDAgents}, 2024.
\newblock Accessed: 2024-06-27.

\bibitem{nemo_guardrails}
NVIDIA.
\newblock Nemo guardrails: A toolkit for building safe and trustworthy conversational agents.
\newblock \url{https://github.com/NVIDIA/NeMo-Guardrails}, 2023.
\newblock Accessed: 2024-10-08.

\bibitem{OpenAI2023GPT4}
OpenAI.
\newblock Gpt-4 system card.
\newblock \url{https://openai.com/research/gpt-4}, 2023.
\newblock Accessed: 2024-06-27.

\bibitem{seamless_communication}
Facebook Research.
\newblock Seamless communication: Multilingual translation and understanding.
\newblock \url{https://github.com/facebookresearch/seamless_communication}, 2023.
\newblock Accessed: 2024-10-08.

\bibitem{tang2023MedAgents}
Xiangru Tang, Anni Zou, Zhuosheng Zhang, Yilun Zhao, Xingyao Zhang, Arman Cohan, and Mark Gerstein.
\newblock Ml-medagents: Large language models as collaborators for zero-shot medical reasoning.
\newblock {\em arXiv preprint arXiv:2311.10537}, 2023.

\bibitem{tchango2022ddxplus}
Arsene~Fansi Tchango, Rishab Goel, Zhi Wen, Julien Martel, and Joumana Ghosn.
\newblock Ddxplus: A new dataset for automatic medical diagnosis, 2022.
\newblock NeurIPS 2022 Datasets and Benchmarks Track, Camera ready.

\end{thebibliography}
\end{document}